\documentclass[sigplan,nonacm]{acmart}
\pdfoutput=1

\usepackage{microtype}
\usepackage{graphicx}
\usepackage{booktabs} 
\usepackage{algorithm}
\usepackage{algpseudocode}
\usepackage{booktabs}
\usepackage{caption}
\usepackage{subcaption}
\usepackage{seqsplit}

\usepackage{hyperref}

\usepackage{xspace}
\usepackage{multirow}

\newcommand{\sysname}{COMM-RAND\xspace}

\author{Vignesh Balaji}
\authornote{Corresponding author: vbalaji@nvidia.com}
\affiliation{
  \institution{NVIDIA}
  \city{}
  \country{USA}
}

\author{Christos Kozyrakis}
\affiliation{
  \institution{NVIDIA \& Stanford University}
  \city{}
  \country{USA}
}

\author{Gal Chechik}
\affiliation{
  \institution{NVIDIA \& Bar-Ilan University}
  \city{}
  \country{Israel}
}

\author{Haggai Maron}
\affiliation{
  \institution{NVIDIA \& Technion University}
  \city{}
  \country{Israel}
}

\begin{document}

\title{Efficient GNN Training Through Structure-Aware Randomized Mini-batching} 

\begin{abstract}

Graph Neural Networks (GNNs) enable learning on real-world graphs and mini-batch training has emerged as the de facto standard for training GNNs because it can scale to very large graphs and improve convergence. 
Current mini-batch construction policies largely ignore efficiency considerations of GNN training.
Specifically, existing mini-batching techniques employ randomization schemes to improve accuracy and convergence. 
However, these randomization schemes are often agnostic to the structural properties of the graph (for eg. community structure), resulting in highly irregular memory access patterns during GNN training that make sub-optimal use of on-chip GPU caches. 
On the other hand, while deterministic mini-batching based solely on graph structure delivers fast runtime performance, the lack of randomness compromises both the final model accuracy and training convergence speed.
In this paper, we present \emph{\underline{Comm}unity-structure-aware \underline{Rand}omized Mini-batching} (\sysname), a novel methodology that bridges the gap between the above extremes.
\sysname allows practitioners to explore the space between pure randomness and pure graph structural awareness during mini-batch construction, leading to significantly more efficient GNN training with similar accuracy.
We evaluated \sysname across four popular graph-learning benchmarks. \sysname cuts down GNN training time by up to 2.76x (1.8x on average) while achieving an accuracy that is within 1.79\% points (0.42\% on average) compared to popular random mini-batching approaches.   
\end{abstract}

\maketitle

\section{Introduction} \label{sec:intro}

Graphs are a primary type of data in many important applications ranging from social network analysis~\cite{gnn-socialrec} to financial fraud detection~\cite{gnn-frauddetection} with typical graphs containing millions of nodes and billions of edges. Graph neural networks (GNNs) have emerged as the primary tool for analyzing such graphs, but the scale of inputs poses significant computational challenges as the graph data far exceeds available memory capacity on GPUs. A common approach to addressing this limitation of GNN training is {\em mini-batching}, where the graph is divided into smaller chunks on which the model is then trained sequentially.


GNN training via mini-batching faces two competing requirements. Randomization during training is essential for model accuracy and convergence \cite{bottou2009curiously, bengio2012practical, safran2020good, gurbuzbalaban2021random}, leading to widespread adoption of random sampling in GNN frameworks \cite{hamilton2017inductive, fey2019fast, wang2019deep}. However, graph analytics research has shown that exploiting graph structure through node reordering can significantly improve cache efficiency \cite{gorder,rabbit,cagra,when-is-reordering-opt,rabbitpp-ispass2023,grasp-hpca20,nested-dissection,parallel-rcm}. This is particularly effective for real-world graphs that exhibit community structure, where nodes are densely connected within communities but sparsely connected across communities. While this community-aware approach has proven effective for GNN inference \cite{grow-hpca23,gnnadvisor-osdi21}, its benefits are lost during training due to random sampling that harms locality.
The result is a clear trade-off: random mini-batching achieves good accuracy and convergence but with slow per-epoch computation, while structure-based mini-batching enables fast per-epoch computation but leads to slower convergence and reduced accuracy. This creates a significant performance dilemma for practitioners.


We present \sysname (Community-Structure-aware Randomized Mini-batching), a GNN training methodology to bridge this gap by introducing both structural bias and randomization into the mini-batching process in a controlled manner.
There are two main steps of mini-batch construction: (1) partitioning the nodes in the training set across batches and (2) building a sub-graph by sampling a subset of the neighborhood of nodes within each batch.
Conventional mini-batch construction uses uniform (or structure-agnostic) randomization for both steps which optimizes for fast convergence in a few epochs at the expense of slow per-epoch processing.
In contrast, we develop a constrained form of randomization that is aware of the community structure in graphs and use it to build mini-batches that are effective at improving per-epoch performance (due to community-bias) while also maintaining good convergence (due to randomization).
Furthermore, \sysname exposes knobs that allow controlling the level of structural-bias and randomness, enabling practitioners to explore the right balance between per-epoch processing and convergence rate to improve overall GNN training time.
In summary, we make the following contributions in this paper:
\begin{itemize}
    \item We highlight the tension between randomization and graph structure awareness during GNN training by showing the pitfalls of both purely-random and purely-structure-based mini-batching approaches (\underline{Section~\ref{sec:rand-vs-structure}})
    \item We introduce {\em community structure-aware randomization} (\sysname) and provide knobs to balance the competing requirements of high per-epoch performance vs model convergence and accuracy (\underline{Section~\ref{sec:biased-rand}})
    \item We perform a design space exploration with \sysname's knobs across multiple datasets and find the knobs that achieve the best balance of performance and accuracy -- an average total training speedup of 1.8x (2.76x max) for an average 0.42\% point (1.79\% max) drop in final validation accuracy (\underline{Section~\ref{subsec:main-knob-sweep}})
    \item We show that with fixed hyperparameter tuning and training budgets, \sysname is able to train for more epochs and improve test accuracy (\underline{Section~\ref{subsec:hyperparam-tuning}})
    \item We show that \sysname offers additional improvements in the presence of a software-managed cache (\underline{Section~\ref{subsubsec:sw-gpu-cache}}) and when the problem size relative to the on-chip cache capacity grows (\underline{Section~\ref{subsec:l2-cache-sensitivity}})
\end{itemize}
\section{Background and prior work} \label{sec:background}

We first provide a background on GNN training and discuss the difference between full-batch vs mini-batch training.
After demonstrating the superiority of mini-batch GNN training, we provide a high-level overview of prior work on mini-batch construction.


{\bf Learning on graphs with GNNs:}
We focus on a common machine learning scenario in which the input consists of a single large graph and goal is to predict the attributes of individual nodes \cite{kipf2016semi}. A typical application is predicting the characteristics of users within a social network.
Consider a graph $G = (V, E)$ with $|V|$ nodes and $|E|$ edges. Its adjacency matrix $A$ has dimensions $|V| \times |V|$. Each node has an $F$-dimensional feature vector, with all node features collected in matrix $X \in R^{|V| \times F}$. In each layer of the graph neural network, a node's embedding is computed using the embeddings of its neighbors from the previous layer using the following update steps:
\begin{equation} \label{eq:gnn-embedding}
X^{l+1} = \sigma(A'X^{l}W^{l}), \quad l=1,\dots,L-1
\end{equation}
Here, $L-1$ is the number of GNN layers in the model, $X^{l+1}$ contains the embeddings of all nodes after layer $l$, $A'$ is the normalized and regularized adjacency matrix, $W^1,\dots, W^{L-1}$ are weight matrices and $\sigma$ is a pointwise non-linearity such as a ReLU function \footnote{Equation \ref{eq:gnn-embedding} is formulated using GCN layers \cite{kipf2016semi} without bias for simplicity, the formulation with other message passing neural networks is similar, see for example \cite{hamilton2017inductive}}.
The goal of training is to learn the layer parameters $W^1,\dots,W^{L-1}$ by minimizing the loss between the labels of all labeled nodes and the node embeddings of the last layer. 
We refer readers to \cite{chiang2019cluster} for more detailed analysis time and space complexity of GNN training methods. 

{\bf Full-batch training:}
One of the earliest GNNs proposed was Graph Convolution Network (GCN) which used full-batch gradient descent for training~\cite{kipf2016semi}, i.e., propagating information through the entire graph simultaneously in every epoch.
Although full-batch training offers fast per-epoch processing times (by virtue of only updating gradients once every epoch and better utilization of GPU resources), full-batch needs to store all the intermediate embeddings for all nodes and layers.
Full-batch training of an $L$ layer GNN with $F$-dimensional feature vectors requires a memory footprint of $O(|V|FL)$ which makes scaling to large graphs infeasible.

\textbf{Mini-batch training:} An alternative to full-batch training is to perform mini-batch gradient descent where the basic unit of computation is a much smaller subset of nodes and the memory footprint reduces to $O(Bd^{L}F)$ where $B$ is the batch size and $d$ is the average degree of the graph.
To further reduce footprint, GraphSAGE~\cite{hamilton2017inductive} pioneered the concept of sampling the neighbors and bounding the maximum number of neighbors used for each node to $r$ further reducing the footprint to $O(Br^{L}F)$.

\textbf{Superiority of mini-batch training:} Even on GPUs with sufficient memory capacity, mini-batch training is still preferable over full-batch training because mini-batch training makes gradient updates multiple times each epoch (once for every batch) and, therefore, converges in fewer epochs.
Our evaluation on the A100 GPU (80GB memory) across the {\tt reddit}, {\tt ogbn-products}, and {\tt igb-small} datasets (Section~\ref{sec:methodology}), shows that mini-batch training requires on average 10.2x fewer epochs to converge than full-batch training.
Consequently, mini-batch training is still 2.7x faster on total training time compared to full-batch training despite the latter having faster per-epoch processing times.
Therefore, beyond improved scalability, mini-batching's randomization enables faster convergence than full-batch training. Hence, we focus on randomized mini-batch training in this work.

{\bf Mini-batches construction techniques:} 
Algorithm~\ref{alg:minibatch} shows a typical mini-batch GNN training pipeline.
Mini-batch construction is a two-step process.
First, we need to divide the nodes in the training set across different batches.
The nodes within each mini-batch are also known as the root nodes and, hence, we refer to this step as root node partitioning (Line 2).
The second step constructs a sub-graph for each batch. 
Constructing the sub-graph involves traversing the $L$-hop neighborhood from all of the root nodes within the batch and then sampling a subset of those neighbors (Line 4).
All root nodes, sampled neighbors, and connections between them form the sub-graph of a specific batch.
The training computations (forward and backward passes) are performed on the sub-graphs (Line 6).
Specifically, the model takes in input features corresponding to the nodes within a batch's subgraph and then creates predictions. These predictions are then compared against the labels (corresponding to the root nodes in the batch) for the backward pass.

\begin{algorithm}[ht]
\caption{Mini-batch Training of an $L$-layer GNN}
\label{alg:minibatch}
\begin{algorithmic}[1]
\For{each training epoch}
    \State {\bf Step-1: Root node partitioning}
    \Statex \qquad Shuffle node IDs in training set $\bigstar$
    \Statex \qquad Divide nodes across mini-batches $\{B_1,...,B_k\}$
    \For{each mini-batch $B_i$}
        \State {\bf Step-2: Sub-graph Construction}
        \Statex \qquad \qquad Find $L$-hop neighbors of root nodes in $B_i$
        \Statex \qquad \qquad Sample a subset of the neighbors $\bigstar$
        \Statex \qquad \qquad Build sub-graph $S_i$ out of root nodes and sampled neighbors
        
    \EndFor
    \State Train on sub-graphs $\{S_1,...,S_k\}$
\EndFor
\end{algorithmic}
\end{algorithm}

GraphSAGE \cite{hamilton2017inductive} popularized the use of uniform randomization for both mapping root nodes across batches and sampling neighbors for the sub-graph construction.
Specifically, GraphSAGE performs uniform random shuffling of nodes in the training set (star under Line-2 in Algorithm~\ref{alg:minibatch}) and uniform random sampling of the $L$-hop neighborhood (star under Line-4).
The rationale for randomization every epoch is to expose the model to a broad range of nodes in the graph and produce diverse mini-batches that connect disparate regions of the graph for better generalization and convergence.
However, a problem with GraphSAGE is that the number of neighbors sampled grows exponentially with each extra layer (known as the {\em neighborhood explosion problem}).

To address the neighborhood explosion problem, various alternatives to mini-batch construction have been proposed (Liu et al. \cite{liu2021sampling} provide a comprehensive survey of GNN mini-batching techniques).
FastGCN~\cite{chen2018fastgcn}, LADIES~\cite{ladies}, and LABOR~\cite{labor} all use variations of importance sampling to sample fewer neighbors and bound the size of batches' sub-graphs.
ClusterGCN~\cite{chiang2019cluster} first partitions the graph and then builds mini-batches by randomly combining different partitions.
A common theme among all these prior mini-batch construction schemes is the use of randomization for achieving good accuracy and convergence with GNN training.
However, with the exception of clusterGCN, the randomization schemes are agnostic to graph structural properties which limits the ability to improve GNN training speed.
ClusterGCN, while a step in the right direction, suffers from limitations due to poor scalability with small training set sizes.
Section~\ref{subsec:prior-work-comparison} provides a quantitative comparison of the above prior work.
\section{Tension between Randomization vs Structure-Awareness in Mini-Batching} \label{sec:rand-vs-structure}

In the previous section we discussed the use of randomization in mini-batch training of GNNs.
This section explores how graph structure and randomization create a trade-off between per-epoch training speed and overall training convergence (number of epochs until convergence).

%
\begin{figure}[t]
    \centering
    \includegraphics[keepaspectratio,width=0.85\columnwidth]{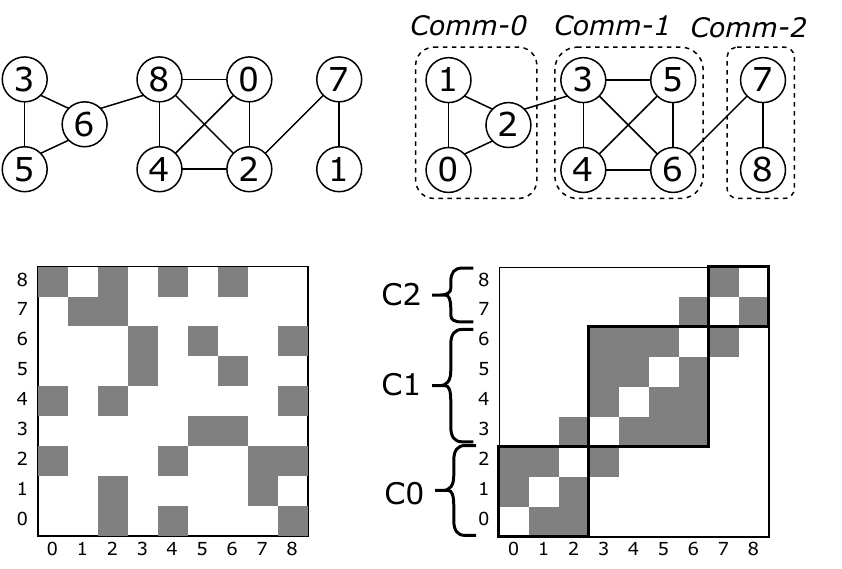}
\caption{{\bf Community-based Graph Reordering:} 
{\em Assigning community members consecutive IDs improves locality}
}
\label{fig:reordering-example}
\end{figure}

{\bf Benefits of community-based graph reordering: } \label{subsec:graph-reordering} 
Real-world graphs exhibit strong structural properties (such as community structure~\cite{communities} and power-law degree distributions~\cite{powerlaw}) that can be leveraged to improve performance on graph-processing applications.
Graph Reordering is one such optimization where the order of nodes in the graph is altered to reflect a graph structural property~\cite{rabbit,when-is-reordering-opt,cagra,rabbitpp-ispass2023,grasp-hpca20,gorder}.
Figure~\ref{fig:reordering-example} shows an example of a graph with random node ordering (left panel).
However, this graph exhibits community structure and reordering the nodes of the graph, such that members of a community have consecutive ordering, significantly improves the sparsity pattern of the graph post-reordering (right panel).

%

Reordering improves performance of graph-based computations by primarily improving data reuse in on-chip caches.
The data footprint of typical graphs often far exceeds the limited cache capacity available in processors.
In an unordered graph (Figure~\ref{fig:reordering-example} left panel), the sparsity pattern leads to a highly irregular memory access pattern that often misses in the cache and requires performing expensive main-memory accesses.
In a community-reordered graph (right panel), the structured sparsity pattern means members of a community are likely to be accessed together which leads to high {\em spatial} and {\em temporal} locality.
Since the data footprint of communities are smaller than the entire graph, community-based reordering improves cache utilization and performance.

{\bf Impact of reordering on GNN inference and training: }
Since GNN inference has a similar memory access pattern as traditional graph analytics workloads and inference accuracy is largely invariant to the node ordering, prior work has applied (community-based) graph reordering to GNN inference and achieved significant reductions in inference time~\cite{grow-hpca23,gnnadvisor-osdi21}.
Our experiments corroborate these findings and we observe that community-based reordering reduces the inference time of a GraphSAGE network by up to 26\% (12\% on average).
However, when applied to GraphSAGE {\em training}, reordering only improves training time by 3\% on average.
The lackluster improvement in GNN training is an artifact of randomization in mini-batch construction.
While essential for improving convergence (Section~\ref{sec:background}), randomization in mini-batching undoes the data reuse advantages of graph reordering\footnote{We note that graph reordering combined with full-batch training could potentially match the performance of standard mini-batching. However, this approach would still face all the limitations of full-batch training.}. 

\begin{figure}[ht]
    \centering
    \begin{subfigure}[]{0.49\columnwidth}
        \centering
        \includegraphics[keepaspectratio,width=\columnwidth]{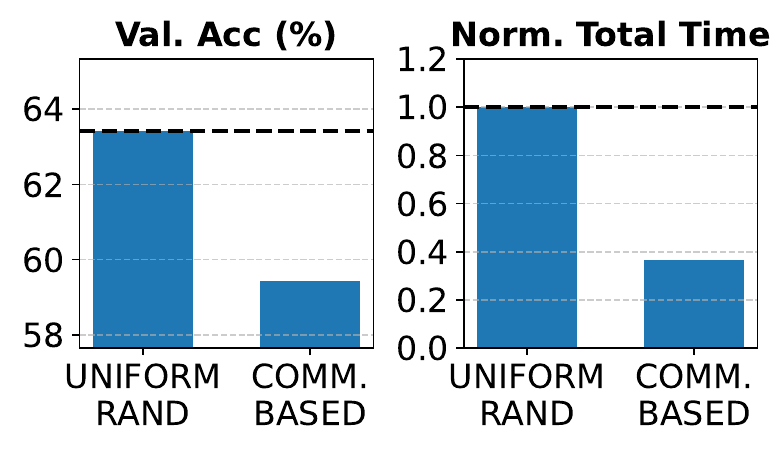}
        \caption{ogbn-papers100M}
        \label{fig:rand-vs-norand-extreme:ogbn-papers100M}
    \end{subfigure}
    \hfill
    \begin{subfigure}[]{0.49\columnwidth}
        \centering
        \includegraphics[keepaspectratio,width=\columnwidth]{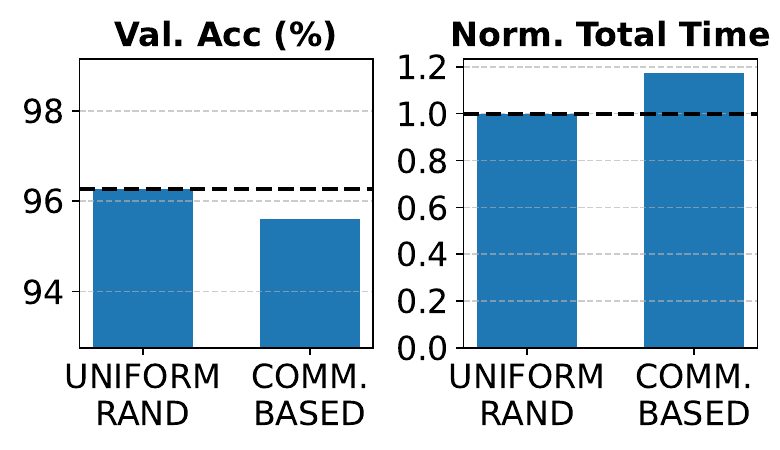}
        \caption{reddit}
        \label{fig:rand-vs-norand-extreme:reddit}
    \end{subfigure}
    \caption{{\bf Impact of entirely community-based mini-batching:}
    {\em Removing randomization leads to (a) accuracy loss and/or (b) a net training slowdown due to delayed convergence}
    }
    \label{fig:rand-vs-norand-extreme}
\end{figure}

{\bf The cost of eliminating randomization: }
The reason why conventional mini-batch training is unable to benefit from graph reordering is because the randomization used during mini-batch construction is agnostic to the graph's structure (i.e. community structure).
Therefore, one possibility for improving mini-batch training performance would be to skip randomization during mini-batch construction.
Specifically, if we do not shuffle node IDs before partitioning the root nodes (Algorithm~\ref{alg:minibatch}; Line 2) and sample the $L$-hop neighborhood (Line 4) in a manner such that we always select neighbors belonging to the same community as the nodes within the batch, then we will have mini-batches that reflect the community structure of a reordered graph.
We compared such an entirely community-based mini-batch construction scheme against the uniform randomization used in conventional mini-batch training and Figure~\ref{fig:rand-vs-norand-extreme} shows results for training a GraphSAGE network on two datasets.
Skipping randomization allows community based mini-batching to benefit from data-reuse improvements of a reordered graph and reduces the per-epoch training time compared to uniform random mini-batching.
For the {\tt ogbn-papers100M} dataset, skipping randomization provides a per-epoch speedup of 4.5x which leads to a net training speedup of 2.7x in spite of taking 1.7x more epochs to converge.
However, entirely community-based mini-batching leads to a significant degradation in final validation accuracy by nearly 4\% points (Figure~\ref{fig:rand-vs-norand-extreme:ogbn-papers100M}).
For the {\tt reddit} dataset, while the accuracy loss is minor, the per-epoch speedup of 1.85x is insufficient to counter a delay in convergence by 2.17x epochs which causes the net training time to {\em increase} by 1.2x (Figure~\ref{fig:rand-vs-norand-extreme:reddit}).
The lack of randomization in an entirely community-based mini-batching results in poor quality mini-batches (low diversity within batches) which impacts convergence and accuracy. 


{\bf Conclusion: }
The above results highlight a fundamental tension between randomization vs leveraging graph structure in mini-batching.
Ignoring graph structure (i.e. uniform random mini-batching) misses an opportunity to optimize per-epoch training time while completely ignoring randomization (entirely community-based mini-batching) comes at the expense of accuracy loss and delayed convergence.


\section{Biased Random Mini-batching} \label{sec:biased-rand}

Uniform random mini-batching and entirely community-based mini-batching represent extreme points on the GNN training convergence vs per-epoch runtime trade-off space. 
The main contribution of our work is to explore the intermediate points of the trade-off space using {\em community structure-aware randomization} (\sysname) for mini-batch construction.
%
%
In this section, we propose community-aware randomization schemes for the two steps of mini-batch construction (Algorithm~\ref{alg:minibatch}) that allows combining the convergence benefits of randomization with the data reuse benefits of structure-awareness.
Our proposal requires the community membership of nodes in the graph which is available to us once we perform community-based graph reordering\footnote{While we use a reordering mechanism based on hierarchical community detection via modularity maximization~\cite{rabbit}, \sysname can work with any community detection algorithm and also does not strictly require the graph to be community-ordered.}.


\subsection{Biased partitioning of root nodes to batches} \label{subsec:bias-root-partitioning}
\begin{figure}[t]
    \centering
    \includegraphics[keepaspectratio,width=0.9\columnwidth]{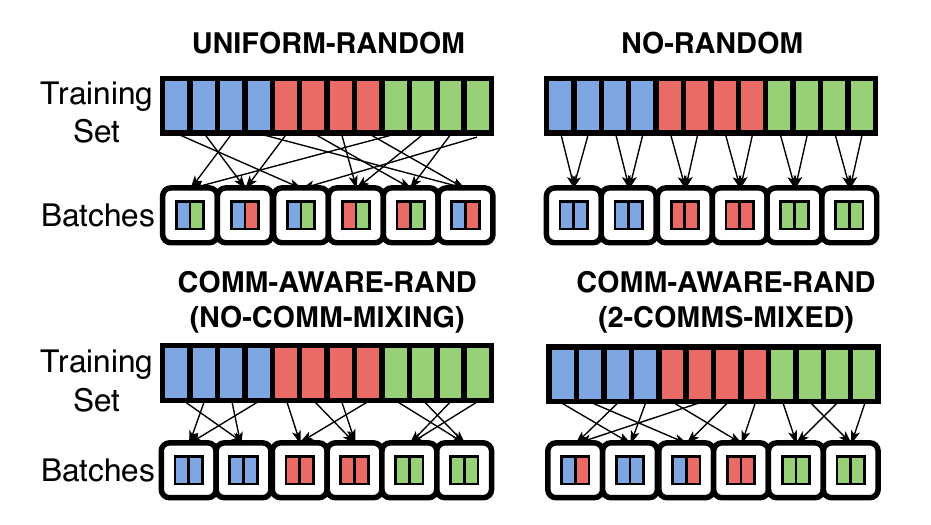}
\caption{{\bf Biased Root Partitioning:} 
{\em Communities in the training set are shown with different colors. For the sake of simplicity, we do not show community-wide randomization for the community-aware schemes in the bottom row. In practice, communities are also shuffled as whole blocks in addition to randomizing the contents within community boundaries.}
}
\label{fig:biased-root-selection}
\end{figure}

The baseline mini-batching approach randomly and uniformly distributes the nodes of the training set across mini-batches.
Specifically, during each training epoch, the contents of the training set are shuffled in a uniform random manner and then the elements of the shuffled set are mapped to batches as demonstrated in the top left panel of Figure~\ref{fig:biased-root-selection}. 
With uniform randomization, the root nodes within each mini-batch often belong to different communities which is sub-optimal for data reuse (as discussed in the previous Section~\ref{sec:rand-vs-structure}, the cache locality benefits of reordering primarily comes from processing members of a community together).
In contrast, the top-right panel of Figure~\ref{fig:biased-root-selection} shows the effect of not randomizing the training set before mapping the root nodes to batches.
In this case, the root nodes within batches are all mostly from the same community and the partitioning of root nodes to batches is static across all training epochs.
While great for data reuse and per-epoch runtime, such a scheme would suffer from poor convergence (Figure~\ref{fig:rand-vs-norand-extreme}).

Aiming to find a good balance between randomization and community awareness in the root node partitioning step, we propose {\em community-aware} randomization of the training set before mapping root nodes to mini-batches (the star under Line-2 of Algorithm~\ref{alg:minibatch} identifies the step we modify).
Specifically, within the training set, we treat each community as a block and then apply two levels of randomization: (1) We shuffle the communities around as whole blocks, and (2) we also shuffle the contents within each community block.
As illustrated in the bottom-left panel of Figure~\ref{fig:biased-root-selection}, this constrained form of randomization results in the root nodes of each mini-batch to mostly belong to the same community (similar to the no randomization policy) while also ensuring that the root nodes of each batch are shuffled every epoch (similar to uniform randomization).

To allow users to vary the level of community structure-bias and randomness during mini-batch creation, we add a knob to the scheme described in the previous paragraph.
Instead of restricting randomization at the granularity of a single community, we allow a fixed number of communities to be {\em mixed} to form a super-community and then restrict randomization to within the super-community boundary. 
Increasing the number of communities that are mixed, increases the amount of randomization and reduces the community-bias during mini-batching and vice versa.
The bottom-right panel of Figure~\ref{fig:biased-root-selection} shows the effect of mixing two communities which creates mini-batches that fall in between the batches of uniform randomization and no randomization.
In order to perform community-aware biased root selection as depicted in the bottom row of Figure~\ref{fig:biased-root-selection}, we only require the community information of each node in the graph which is available to us after performing community-based reordering.
{\bf In summary, the \sysname knob for controlling the community structure bias in partitioning root nodes is the number of communities to mix before randomization (value ranges from 0 to number of communities in the training set).}


\subsection{Biased neighborhood sampling} \label{subsec:biased-ngh-sampling}
\begin{figure}[t]
    \centering
    \includegraphics[keepaspectratio,width=0.9\columnwidth]{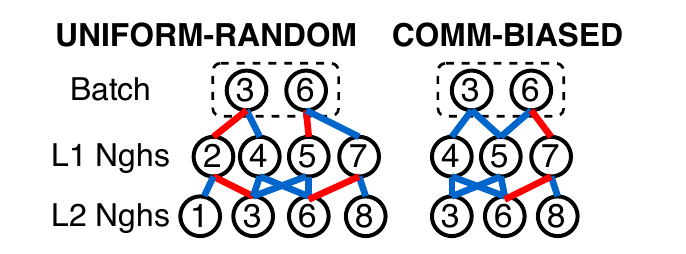}
\caption{{\bf Biased Neighborhood Selection:} 
{\em Biased neighborhood selection for the graph in Figure~\ref{fig:reordering-example}. Intra-community edges are highlighted with blue and inter-community edges are highlighted with red. For a 2-layer GNN, uniform random neighborhood sampling creates a larger sub-graph compared to a biased scheme where intra-community neighbors are selected with a higher probability.}
}
\label{fig:biased-ngh-selection}
\end{figure}

Once the root nodes of each mini-batch are known, the next step is to traverse the $L$-hop neighborhood of the root nodes (for a $L$-layer GNN) and sample a subset of the neighbors to form the sub-graph associated with each batch (star under Line-4 of Algorithm~\ref{alg:minibatch}).
Figure~\ref{fig:biased-ngh-selection} shows the sub-graph created for a two-layer GNN operating on the example reordered graph from Figure~\ref{fig:reordering-example}.
The baseline approach samples these neighbors in a uniform random manner where each neighboring node has an equal likelihood of being selected.
Consequently, the sub-graph with uniform random neighbor sampling consists of nodes spanning multiple communities which increases the batch's data footprint and increases the cache pressure.
A community-biased sampling approach differentiates between {\em intra-} and {\em inter-}community edges and samples neighbors belonging to the same community with a higher probability.
The right-panel of Figure~\ref{fig:biased-ngh-selection} illustrates that prioritizing intra-community edges results in a smaller sub-graph for the same root nodes.
We set the probability of sampling intra-community edges as $p$ and inter-community edges as $1-p$.
{\bf The \sysname knob for controlling the community structure bias during neighborhood sampling is $p$ where values range from 0.5 (equal likelihood of selecting all neighbors) to 1.0 (only select neighbors from the same community).}

%
In this section, we proposed community-aware randomization schemes for both root node partitioning and neighborhood sampling.
Additionally, we exposed knobs to control the level of community structure bias in the two steps.
In Section~\ref{sec:eval}, we will evaluate GNN training across different combinations of knobs for biased root partitioning and biased neighborhood sampling and demonstrate how controlling the level of graph structure bias yields different points on the per-epoch runtime vs convergence trade-off space.



\section{Experimental Methodology} \label{sec:methodology}

In this section, we provide details about the datasets, software, training methodology, and the platform used to evaluate the effectiveness of \sysname.

{\bf Software and implementation: }
We perform all our evaluations on the NVIDIA NGC container~\cite{nvidia-dgl-container} of the Deep Graph Library (DGL)~\cite{wang2019deep} framework (v24.07).
We use DGL's in-built APIs for different neighborhood sampling schemes and extend the reference implementations of GraphSAGE and other GNN models.
Specifically, we implement the different root node partitioning schemes by creating appropriate permutations of the tensor containing the training set before passing it to DGL's DataLoader. 
To implement biased neighborhood sampling, we use DGL's NeighborSampler API that allows specifying the (unnormalized) probability of each edge in the graph~\cite{dgl-ngh-sampling} and then set the probability of all intra-community edges to $p$ and the probability of inter-community edges to $1-p$. 

\textbf{Evaluated schemes.} Table~\ref{table:root-mapping-policies} lists the root node partitioning policies we evaluate. 
For the \sysname knob for biased root partitioning, we represent the number of communities to mix ($k$) as a percentage of the total number of communities in the training set and evaluate 3 values of $k$ (12.5\%, 25\%, and 50\%).
We evaluate three values of $p$ for biased neighborhood sampling: $0.5$ (equal likelihood of sampling all edges in the graph), $0.9$ (the probability to sample a neighbor from the same community is $9$ times higher than a neighbor from another community), and $1.0$ (only sample neighbors from the same community).
\begin{table*}[ht]
\caption{{\bf Root node partitioning schemes evaluated}}
\label{table:root-mapping-policies}
\centering
\resizebox{1.0\textwidth}{!}{%
\begin{tabular}{ll}
\toprule
\textbf{Root Node Partitioning }   
& \textbf{Description}  \\ \midrule
RAND-ROOTS            & Uniform random shuffling of the training set before dividing nodes across batches  \\
NORAND-ROOTS         & No shuffling of the training set before dividing nodes across batches (root node partitioning is static across epochs) \\ 
COMM-RAND-MIX-0\%   & Shuffle communities as whole blocks; then shuffle contents within each community  \\ 
COMM-RAND-MIX-k\% & Shuffle communities as blocks; Combine fixed no. of communities (k\% of \#communities) into super-blocks; Shuffle within each super-block  \\ 
\bottomrule
\end{tabular}%
}
\end{table*}

{\bf Datasets: }
We evaluate \sysname across four graph datasets that are commonly used for evaluating GNNs (Table~\ref{table:datasets}).
The graphs used in our evaluation are diverse across multiple dimensions -- domains (social, citation, and product-purchasing networks), graph size (232K to 111M nodes), feature dimensions (100 to 1024), and training split (1\% to 65\% of nodes labeled).
For most of our evaluation, we assume that the graphs are already community-ordered because, in addition to powering \sysname, community-based reordering is an effective optimization for GNN inference as well (Section~\ref{sec:background}).
Only the uniform random mini-batching baseline for ClusterGCN comparisons (Section~\ref{subsec:prior-work-comparison}) runs on the original graph ordering.
We use RABBIT~\cite{rabbit} to perform community-based graph reordering and find the community membership of each node.
Section~\ref{subsec:preprocessing-cost} shows that reordering would impose negligible overhead.
As mentioned in Appendix B.4 of the OGB paper~\cite{ogb}, we use the symmetrized version of the {\tt ogbn-papers100M} dataset since the directed version leads to poor accuracy with GraphSAGE.

\begin{table}[h]
    \caption{{\bf Graph Datasets}}
    \label{table:datasets}
    \centering
    \setlength{\tabcolsep}{2pt} %
    \resizebox{\columnwidth}{!}{%
    \begin{tabular}{lccccc}
    \toprule
    \textbf{Graph}  & \textbf{\#Nodes} & \textbf{\#Edges} & \textbf{\#Labels} & \textbf{\#Feat} & \textbf{Train-Val-Test \%} \\ \midrule
    reddit~\cite{reddit-dgl-dataset}          & 232,965          & 114,615,892      & 41                & 602             & 66-10-24                \\ 
    igb-small~\cite{igb}       & 1,000,000        & 13,070,502       & 19                & 1024            & 60-20-20                \\ 
    ogbn-products~\cite{ogb}   & 2,449,029        & 123,718,280      & 47                & 100             & 8-2-90                  \\ 
    ogbn-papers100M~\cite{ogb} & 111,059,956      & 3,228,124,712    & 172               & 128             & 1.1-0.1-0.2             \\ 
    \bottomrule
    \end{tabular}%
    }
\end{table}

{\bf Training methodology:}
Most of our evaluation focuses on training a 3-layer GraphSAGE model (we evaluate \sysname's impact on other GNN models in Section~\ref{subsec:other-gnn-models}).
For our evaluations, we perform training of up to 100 epochs (50 epochs for {\tt ogbn-papers100M} due to its size) with early-stopping.
We stop training if the validation loss does not improve for 6 epochs and measure the final validation accuracy after convergence.
Additionally, we also use PyTorch's {\em ReduceLROnPlateau} learning rate scheduler with default parameters and a patience of 3 epochs.
Training is performed on the default hyper-parameters used in the DGL's reference implementation of GraphSAGE (Batchsize=1024, fanout=10, learning rate=1e-3, weight decay=5e-4, and hidden dim=256). 
We calculate training metrics by averaging results across 5 random seeds (the same seeds are used for all configurations).


{\bf Hardware platform: }
We perform our experiments on an NVIDIA Ampere A100 GPU with 80GB memory, 40MB of L2 cache, and 2039 GB/s of peak memory bandwidth.
A100's memory capacity allows us to store the graph and features on the GPU device memory for {\tt reddit}, {\tt ogbn-products}, and {\tt igb-small} datasets and, therefore, the training on these datasets is done purely on the GPU.
For {\tt ogbn-papers100M}, the graph and feature data are stored in pinned host (CPU) memory and DGL's dataloader uses Unified Virtual Addressing (UVA)~\cite{dgl-dataloader-api} to perform mixed CPU-GPU training.
\section{Evaluation} \label{sec:eval}

We perform an extensive analysis of the impact of \sysname based mini-batching on GNN training by sweeping different combinations of the knobs presented in Section~\ref{sec:biased-rand}.
Specifically, we characterize how changing the level of graph structure bias affects various training metrics   .
We also compare \sysname against prior work on mini-batch construction for GNNs and measure \sysname's sensitivity to software caches for reducing CPU-GPU communication, L2 cache capacity, and hyper-parameter tuning.
%

\subsection{Quantitative analysis of biased mini-batching} \label{subsec:main-knob-sweep}

\begin{figure*}[h!]
    \centering
    \includegraphics[keepaspectratio,width=1.0\textwidth]{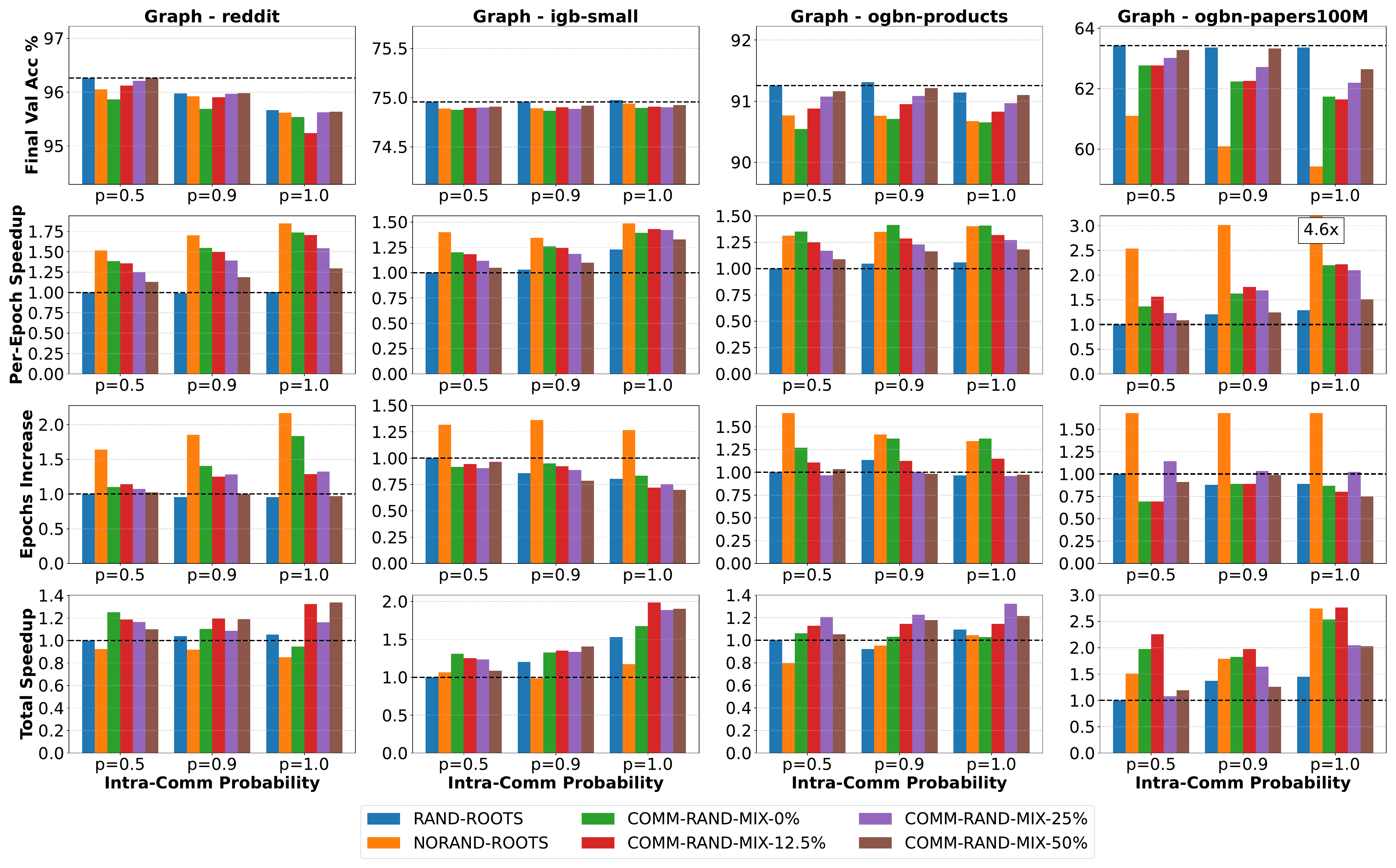}
    \caption{{\bf Impact of \sysname across different GNN training metrics (rows) and input graphs (columns):}
    {\em All results (besides accuracy) are normalized to the baseline ({\tt RAND-ROOTS} \& $p=0.5$). The y-axis for val. acc. does not start from 0. } 
    }
    \label{fig:metrics-vs-knobs}
\end{figure*}

We evaluate the effectiveness of different levels of community biased mini-batching on four training metrics -- final validation accuracy, per-epoch training time, number of epochs until convergence, and total training time.
The rows of Figure~\ref{fig:metrics-vs-knobs} correspond to each of the above metrics and the columns of the figure show results across different graphs.
Within each subplot, the different bars/colors represent different root node partitioning schemes (Table~\ref{table:root-mapping-policies}) and the x-axis represents different values of the intra-community probability knob from Section~\ref{subsec:biased-ngh-sampling}.
In Figure~\ref{fig:metrics-vs-knobs}, the setting ({\tt RAND-ROOTS} \& $p=0.5$) represents uniform random mini-batching baseline and ({\tt NORAND-ROOTS} \& $p=1.0$) represents the entirely community-based mini-batching scheme from Section~\ref{sec:rand-vs-structure}.
Other combinations of root node partitioning schemes and $p$ values represent \sysname with varying levels of community bias during mini-batching.




We make two broad observations across the results in Figure~\ref{fig:metrics-vs-knobs}.
First, while the level of graph structure bias affects the final accuracy, the validation accuracy is always within 1.79\% point of the baseline (with the exception of entirely community-based mini-batching for {\tt ogbn-papers100M} which incurs a 4\% point accuracy loss likely due to its shorter maximum epoch limit). 
Second, we often observe a trade-off between per-epoch training speedup and the number of epochs until convergence.
Over the next few subsections, we will characterize how different levels of community structure bias in \sysname mini-batching impact the per-epoch time and convergence (in turn, affecting the overall speedup).

\subsubsection{Impact on Per-Epoch Training Time} \label{subsec:per-epoch-explanation}
The second row in Figure~\ref{fig:metrics-vs-knobs} reports per-epoch training speedups relative to the baseline of uniform random mini-batching ({\tt RAND-ROOTS} \& $p=0.5$).

\textbf{The impact of root node partitioning:} We first discuss the impact of different root node partitioning policies on per-epoch times (and focus on the results with no intra-community bias during neighborhood sampling $p=0.5$).
As mentioned in Section~\ref{sec:methodology}, we use community-reordered inputs for all our evaluations.
Therefore, the highest per-epoch speedups are achieved when no randomization is performed before partitioning the root nodes across batches ({\tt NORAND-ROOTS}) because the graph's ordering is optimized for high data reuse from caches (Section~\ref{subsec:graph-reordering}).
On average, ({\tt NORAND-ROOTS} \& $p=0.5$) offers 1.69x per-epoch speedups over the baseline.
Community-aware randomization without any community mixing ({\tt COMM-RAND-MIX-0\%}) often achieves the next best per-epoch speedups (1.32x on average) because it creates mini-batches with root nodes that mostly belong to the same community but deviates slightly from the reuse-optimized ordering. 
%
%
Once we start mixing communities ({\tt COMM-RAND-MIX-\{12.5\%,25\%,50\%\}}), we see a further reduction in per-epoch speedups (with speedups decreasing as more communities are mixed).
Even with the \sysname version with the least amount of community structure bias ({\tt COMM-RAND-MIX-50\%}), we notice an average per-epoch speedups of 1.09x over the baseline because the memory access pattern still has a small level of community bias that improves data reuse in caches.

\textbf{The impact of intra-community bias during neighborhood sampling ($p$):} 
Increasing $p$ has a straightforward impact on per-epoch performance -- as the likelihood of selecting intra-community edges increases, the per-epoch speedup also increases because the sub-graphs associated with batches are more aligned with the community structure of the graph.
Furthermore, increasing the intra-community bias during neighbor selection tends to improve per-epoch speedups across {\em all} root node partitioning policies.

\begin{figure}[ht]
    \centering    \includegraphics[keepaspectratio,width=\columnwidth]{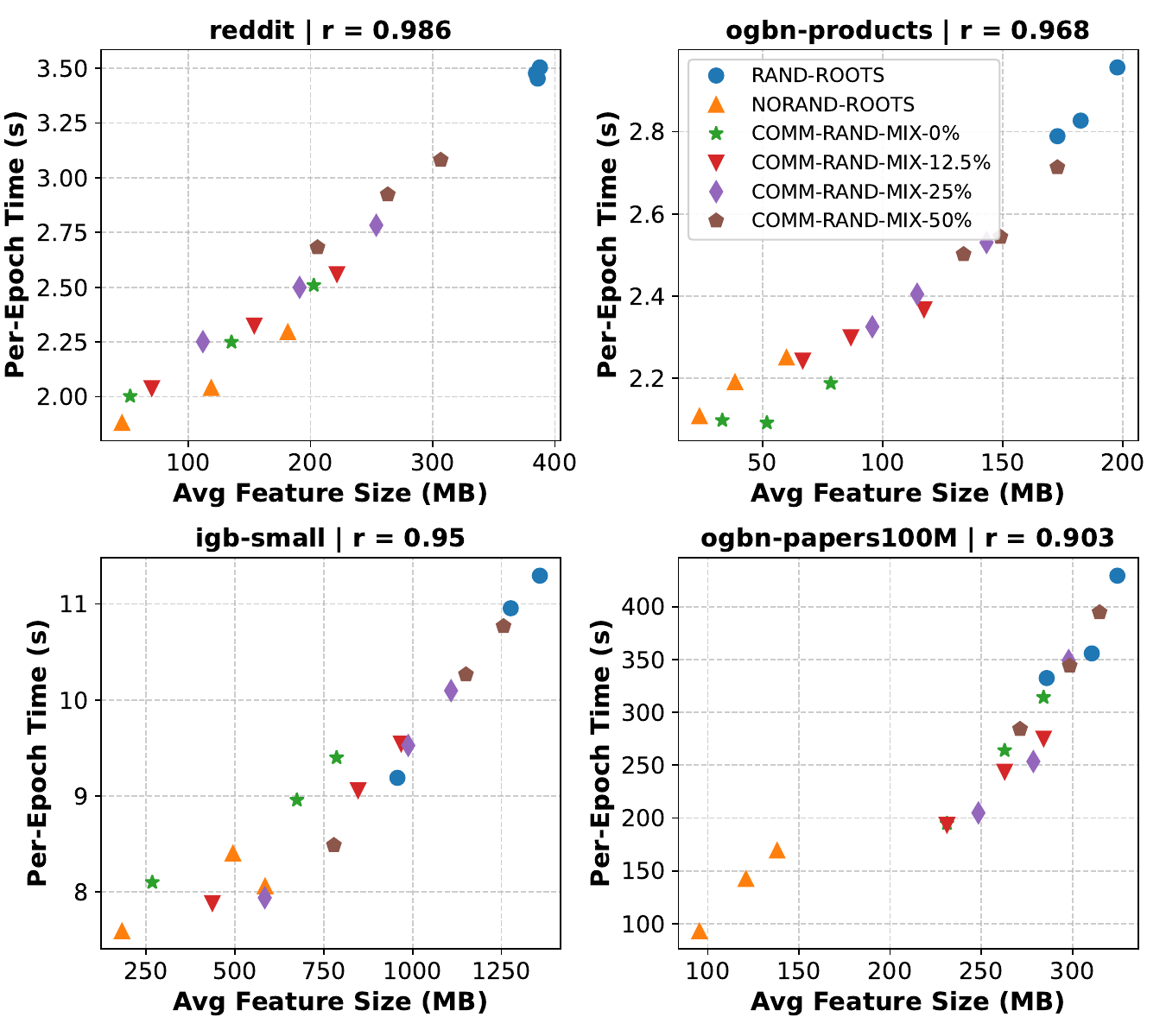}
    \caption{{\bf Per-epoch training time vs input node features size:} 
    {\em Different points of the same color represent different levels of community bias during neighborhood sampling ($p$) for the same root node partitioning policy. The titles also report the pearson correlation between per-epoch time and input feature size for each graph. } 
    }
    \label{fig:perepochtime-vs-featsize}
\end{figure}

\textbf{Discussion:} Results indicate that \sysname mini-batching primarily improves per-epoch training time by reducing the size of the sub-graphs per batch which reduces the footprint of the input feature data that needs to be fetched for the training computations.
To corroborate this, Figure~\ref{fig:perepochtime-vs-featsize} shows the correlation between per-epoch time and the average size of the input feature data ($X$) corresponding to the nodes within batches.
Across root node partitioning policies, we see that the average input node features size decreases as the level of community-bias increases from {\tt RAND-ROOTS} to {\tt COMM-RAND-MIX-50/25/12.5\%} to {\tt COMM-RAND-MIX-0\%} to {\tt NORAND-ROOTS} root node partitioning policies.
Within the same root node partitioning policy, we also see that increasing levels of intra-community bias during neighbor sampling (higher $p$) also reduces the input feature size.
The smaller input feature sizes across batches translate to higher data reuse in software and on-chip caches as demonstrated in Sections~\ref{subsubsec:sw-gpu-cache} and ~\ref{subsec:l2-cache-sensitivity} respectively.


\subsubsection{Impact on convergence}
The third row in Figure~\ref{fig:metrics-vs-knobs} shows the increase in number of epochs until convergence relative to the baseline (lower values are desirable).
In the previous subsection, we noticed that as the level of community-bias in \sysname mini-batching increased, the per-epoch speedups also improved. 
%
%
We observe nearly the opposite trend for convergence -- higher levels of community-bias in \sysname lead to greater delays in convergence.
%
For biased neighborhood sampling with $p=1.0$, the average increase in the number of epochs until convergence (relative to uniform random baseline) is 1.23x, 0.99x, 1.01x, and 0.85x with {\tt COMM-RAND-MIX-0\%}, {\tt COMM-RAND-MIX-12.5\%}, {\tt COMM-RAND-MIX-25\%}, and {\tt COMM-RAND-MIX-50\%} respectively. 

\begin{figure}[ht]
    \centering    \includegraphics[keepaspectratio,width=\columnwidth]{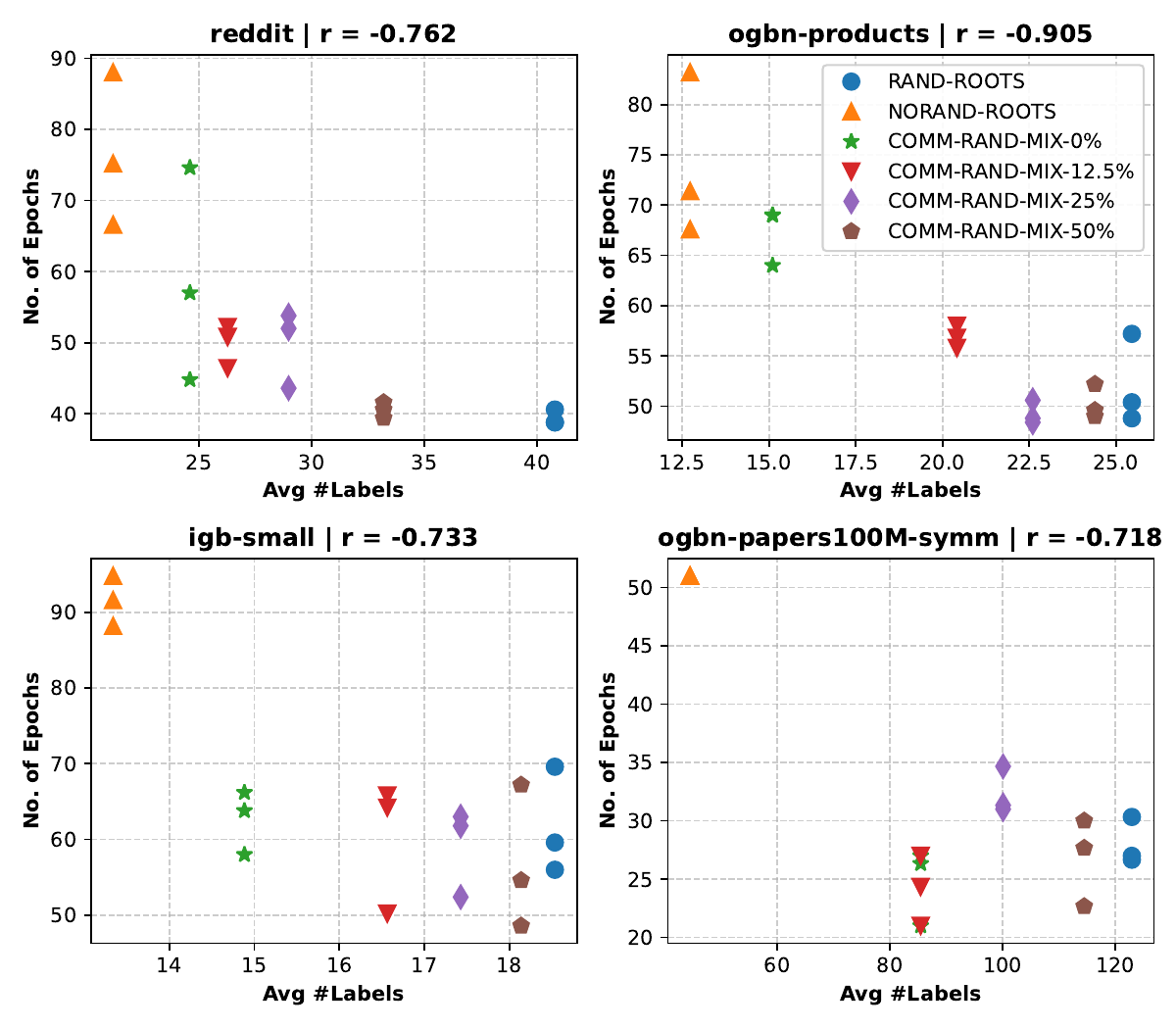}
    \caption{{\bf No. of epochs to converge vs label diversity:} 
    {\em A higher average no. of labels per batch is correlated to faster convergence. Note that the labels are only associated with root nodes and, therefore, different levels of community bias in neighborhood sampling ($p$) has no impact on the label count}
    }
    \label{fig:epochs-vs-numlabels}
\end{figure}

%
Prior work~\cite{chiang2019cluster} observed that the convergence of GNN training is improved when the label entropy among batches is high.
Figure~\ref{fig:epochs-vs-numlabels} reports the relationship between number of epochs until convergence and the average number of labels per batch.
While the correlations are not as strong as Figure~\ref{fig:perepochtime-vs-featsize}, a higher label diversity within batches is correlated with lower number of epochs until convergence.
The result shows that as the level of community bias increases (from {\tt RAND-ROOTS} to {\tt NORAND-ROOTS}), the average number of labels per batch decreases which leads to delayed convergence.

\subsubsection{Best \sysname knobs}

The previous subsections showed a trade-off between per-epoch training time and the number of epochs until convergence.
As the level of community-bias in \sysname mini-batching increases, we see higher per-epoch speedups and greater delay in convergence and vice versa.
Total training time is a product of per-epoch time and the number of epochs. 
Therefore, the last row of Figure~\ref{fig:metrics-vs-knobs} has less obvious trends across different root node partitioning policies and intra-community probabilities ($p$).
However, our evaluation shows that root node partition schemes with community mixing combined with maximum intra-community bias during neighbor sampling ($p=1.0$) provides a good balance between per-epoch speedup and convergence.
Across all datasets, we find that {\tt COMM-RAND-MIX-12.5\%} $+$ $p=1.0$ provides the higher average total training speedups of 1.8x (with speedups ranging from 1.14x to 2.76x).
Selecting other root node partitioning policies does not result in a major performance cliff -- {\tt \seqsplit{COMM-RAND-MIX-25\%}} leads to an average net speedup of 1.6x (range from 1.16x to 2.05x) and {\tt COMM-RAND-MIX-50\%} leads to an average net speedup of 1.62x (range from 1.21x to 2.03x).

The above discussion on the best \sysname knob configuration assumed a scenario where the GNN is trained until convergence.
If instead, the GNN is trained for a fixed number of epochs, then a higher level of community-bias in \sysname mini-batching would be more desirable.
Finally, we note that the knobs exposed in Section~\ref{sec:biased-rand} can allow users to find the right level of community bias for their unique combinations of dataset, GNN model, and GPU architecture. It may even be possible to cast the problem of finding the right bias level as a learning problem in itself and we leave this exploration for future work. 

\subsection{Impact of hyperparameter tuning} \label{subsec:hyperparam-tuning}

The results so far were collected using the default hyper-parameters from the DGL reference implementations.
\sysname introduces two new hyper-parameters -- root node partitioning policy and community bias during neighborhood sampling (as described in Section~\ref{sec:biased-rand}).
A natural question is whether the additional cost of hyperparameter tuning with two extra parameters would overshadow \sysname's benefits showcased in the previous section.
To answer this question, we performed an experiment with fixed budgets for hyperparameter tuning and training using the {\tt reddit} dataset.
For both the baseline and \sysname, we perform a random search for 1 hour (we sweep the 5 hyper-parameters mentioned in Section~\ref{sec:methodology} for the baseline and 7 hyper-parameters for \sysname).
Due to its higher performance, \sysname explores 70 hyper-parameter combinations compared to 62 combinations with the baseline.
Next we trained both the baseline and \sysname versions with the best hyper-parameters for 30 minutes and Table~\ref{table:hyperparam-sweep} reports the results.
Due to superior per-epoch performance, \sysname is able to train for 1.54x more epochs compared to the baseline within the same training budget.
Consequently, \sysname trains a model which achieves a higher final validation accuracy and also improves the inference accuracy by 0.27\% points.
This result demonstrates that \sysname is able to offer benefits even after accounting for the hyperparameter tuning costs.

\begin{table}[ht]
    \caption{{\bf \sysname's effectiveness after hyperparameter tuning:}
    {\em Results after training for 30 minutes (using the best hyper-parameters). \sysname trains for more epochs leading to improved model accuracy}
    }
    \label{table:hyperparam-sweep}
    \centering
    \resizebox{\columnwidth}{!}{%
    \begin{tabular}{lccc}
    \hline
                       & \multicolumn{1}{l}{\textbf{No. of Epochs}} & \multicolumn{1}{l}{\textbf{Final Val. Acc}} & \multicolumn{1}{l}{\textbf{Test Acc}} \\ \hline
    \textbf{Baseline}  & 641.8                                      & 96.10\%                                     & 96.30\%                               \\ 
    \textbf{COMM-RAND} & 987.6                                      & 96.22\%                                     & 96.57\%                               \\ \hline
    \end{tabular}%
    }
\end{table}

\subsection{Comparison to prior work on mini-batching} \label{subsec:prior-work-comparison}

{\bf Structure-agnostic randomized mini-batching: }
As discussed in Section~\ref{sec:background}, most prior-work on mini-batch construction ignores the graph structure.
We choose LABOR as a representative among this class of mini-batching policies because it was shown to outperform LADIES~\cite{labor} (which, in turn, outperforms FastGCN~\cite{ladies}).
After 25 epochs of training on the {\tt reddit} dataset, LABOR achieved an accuracy of 96.08\% (compared to 96.21\% with the uniform random mini-batching baseline) while improving the per-epoch processing time by 1.1x compared to the baseline.
In comparison, \sysname achieves an accuracy of 95.25\% while providing a higher per-epoch speedup of 1.75x.
LABOR (and other related mini-batching schemes) do not take into account the community structure of graphs and, therefore, miss out on the opportunity to significantly improve data reuse.


{\bf Structure-aware randomized mini-batching: }
ClusterGCN~\cite{chiang2019cluster} is the prior work most closely related to ours because they also combine randomization and graph structure while creating mini-batches.
Specifically, ClusterGCN uses METIS~\cite{metis} to partition the graph and then forms mini-batches by randomly combining a fixed number of graph partitions.
Both graph partitioning and community-detection (used by \sysname) find regions of high inter-connectivity within graphs but differ in the same way as k-means clustering differs from agglomerative clustering.
Graph partitioning requires the user to specify the number of partitions whereas community-detection determines the appropriate number of communities based on the graph structure.
However, the differences between ClusterGCN and our work (\sysname) run deeper than the choice of using graph partitioning versus community-based reordering.

\begin{table}[ht]
\caption{{\bf Comparison against clusterGCN across datasets}:
{\em Results after 25 epochs of training}
}
\label{table:clustergcn-comparison}
\centering
\resizebox{\columnwidth}{!}{%
\begin{tabular}{lcccccc}
\hline
              & \multicolumn{2}{c}{\textbf{Baseline}}                                 & \multicolumn{2}{c}{\textbf{COMM-RAND}}                                & \multicolumn{2}{c}{\textbf{ClusterGCN}}                               \\ 
              \midrule 
              & \multicolumn{1}{l}{Per-Epoch } 
              & \multicolumn{1}{l}{Val.  \%} & \multicolumn{1}{l}{Per-Epoch } & \multicolumn{1}{l}{Val. \%} & \multicolumn{1}{l}{Per-Epoch } & \multicolumn{1}{l}{Val. \%} \\ 
              & \multicolumn{1}{l}{ Speedup} 
              & \multicolumn{1}{l}{ Acc \%} 
              & \multicolumn{1}{l}{  Speedup} & \multicolumn{1}{l}{Acc \%} 
              & \multicolumn{1}{l}{  Speedup} & \multicolumn{1}{l}{Acc \%} \\ 
              
              \midrule 
              
reddit        & \multicolumn{1}{c}{1.0x}      & 96.13                            & \multicolumn{1}{c}{1.76x}           & 95.25                            & \multicolumn{1}{c}{2.10x}           & 93.45                            \\ 
igb-small     & \multicolumn{1}{c}{1.0x}          & 74.77                            & \multicolumn{1}{c}{1.68x}          & 74.64                            & \multicolumn{1}{c}{5.94x}              & 74.58                            \\ 
ogbn-products & \multicolumn{1}{c}{1.0x}           & 91.02                            & \multicolumn{1}{c}{2.04x}           & 89.87                            & \multicolumn{1}{c}{0.26x}          & 89.4                             \\ 
ogbn-papers100M   & \multicolumn{1}{c}{1.0x}          & 63.29                            & \multicolumn{1}{c}{1.48x}          & 62.15                            & \multicolumn{1}{c}{0.08x}         & 50.78                            \\ \bottomrule
\end{tabular}%
}
\end{table}

\begin{figure}[ht]
\centering
\includegraphics[keepaspectratio,width=0.9\columnwidth]{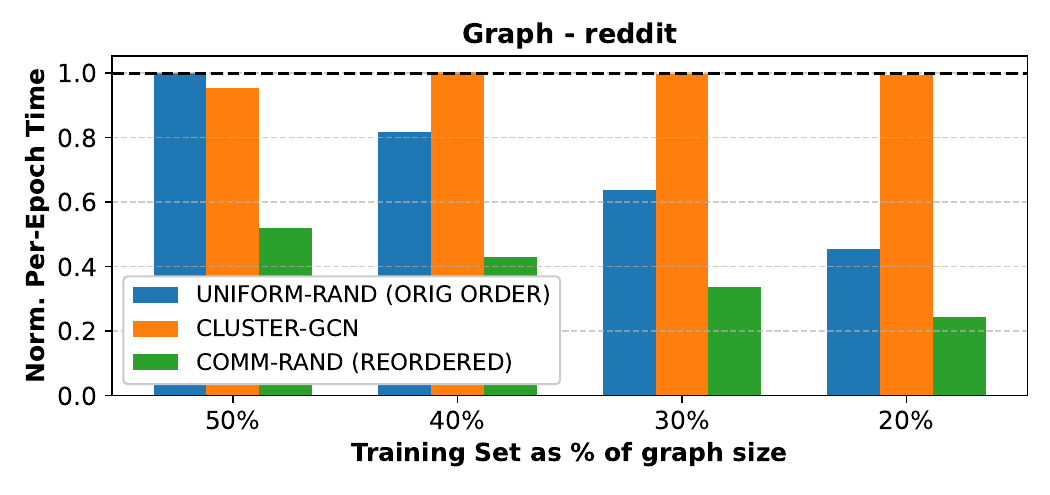}
\caption{{\bf Per-Epoch time vs training set size:}
{\em ClusterGCN performance is invariant to the training set size}
}
\label{fig:perepoch-vs-trainidx-sweep}
\end{figure}

There are two fundamental differences between ClusterGCN and our proposal of \sysname mini-batching.
First, ClusterGCN performs a more limited form of randomization compared to our proposal.
While ClusterGCN creates mini-batches by randomly combining different graph partitions, the contents within each partition are not randomized.
In contrast, our community-aware randomization schemes shuffle entire communities as well as randomizing the contents within each community (Figure~\ref{fig:biased-root-selection}).
Consequently, ClusterGCN converges more slowly in comparison to our proposal and consistently achieves lower accuracy compared to \sysname (Table~\ref{table:clustergcn-comparison}).
Second, ClusterGCN creates mini-batches by composing entire graph partitions and does not take the training set into account during mini-batch construction.
Therefore, ClusterGCN is required to evaluate the {\em entire} graph during training.
For graphs with large training sets ({\tt reddit} \& {\tt igb-small} have 65\% and 60\% of the graph's nodes labeled respectively), ClusterGCN is effective at improving per-epoch time over the baseline (Table~\ref{table:clustergcn-comparison}).
However, when the training set is small ({\tt ogbn-products} and {\tt ogbn-papers100M} have only 20\% and 1\% of the graph's nodes labeled), ClusterGCN is significantly slower than the baseline (by up to 13x).
In contrast, \sysname offers consistent speedups because we only perform computations on the subset of the graph connected to the nodes in the training set.
Figure~\ref{fig:perepoch-vs-trainidx-sweep} further illustrates this inefficiency of ClusterGCN.
When the training set size of {\tt reddit} is artificially reduced, we observe that the per-epoch training time with ClusterGCN remains constant while both the baseline and \sysname mini-batching offer lower per-epoch processing times as the training set size shrinks.
Therefore, ClusterGCN is unlikely to be effective in the more realistic GNN training scenario where only a small percentage of the graph's nodes are labeled.

\subsection{Generalizing to other GNN models} \label{subsec:other-gnn-models}

The results up to this point have focused on the GraphSAGE model. 
However, \sysname generalizes to arbitrary GNN models.
To illustrate this point, we performed a sweep of different \sysname knobs (as performed in Figure~\ref{fig:metrics-vs-knobs}) on two other popular GNN models -- Graph Convolution Network (GCN)~\cite{kipf2016semi} and Graph Attention Network (GAT)~\cite{velivckovic2017graph}.
Table~\ref{table:other-gnn-models} shows different training metrics for the {\tt reddit} dataset where the baseline is {\tt RAND-ROOTS} $+$ $p = 0.5$ and \sysname reports results for the knobs that achieve the best balance between accuracy and performance.
The results show that for both the GCN and GAT, \sysname achieves a final validation accuracy that is within 1\% points of the baseline while improving both the per-epoch and total training times.
\sysname achieves a total training speedup of 2.03x and 1.38x for GCN and GAT respectively.

%


\begin{table}[ht]
    \caption{{\bf \sysname's effectiveness for other GNN models:}
    {\em Results for the {\tt reddit} dataset.}
    }
    \label{table:other-gnn-models}
    \centering
    \resizebox{\columnwidth}{!}{%
    \begin{tabular}{llcccc}
    \toprule
    &   & \multicolumn{1}{c}{Final Val.} & \multicolumn{1}{c}{Per-Epoch} & \multicolumn{1}{c}{ Num} & \multicolumn{1}{c}{Total Training } \\ 
    &   & \multicolumn{1}{c}{Acc. \%} & \multicolumn{1}{c}{Time (s)} & \multicolumn{1}{l}{Epochs} & \multicolumn{1}{c}{Time (s)} \\ 
    \midrule
    \multirow{2}{*}{\begin{tabular}[c]{@{}l@{}} {\bf GCN} \end{tabular}}   & Baseline  & 78.67 & 3.42   & 50.4   & 172.18    \\  
    \cline{2-6} 
    & COMM-RAND & 77.82   & 2.86    & 29.6   & 84.56   \\ \midrule
    \multirow{2}{*}{\begin{tabular}[c]{@{}l@{}} {\bf GAT} \end{tabular}} & Baseline  & 90.60                                  & 2.00                                    & 51.0                               & 102.05                                       \\ \cline{2-6} 
         & COMM-RAND & 90.09                                  & 1.74                                    & 42.4                               & 73.72                                       \\ \bottomrule
    \end{tabular}%
    }
    \vspace{-1.5ex}
\end{table}


\subsection{Sensitivity studies}
\subsubsection{Impact of software-managed caches.} \label{subsubsec:sw-gpu-cache}
\begin{figure}[h]
\centering
\includegraphics[keepaspectratio,width=1.0\columnwidth]{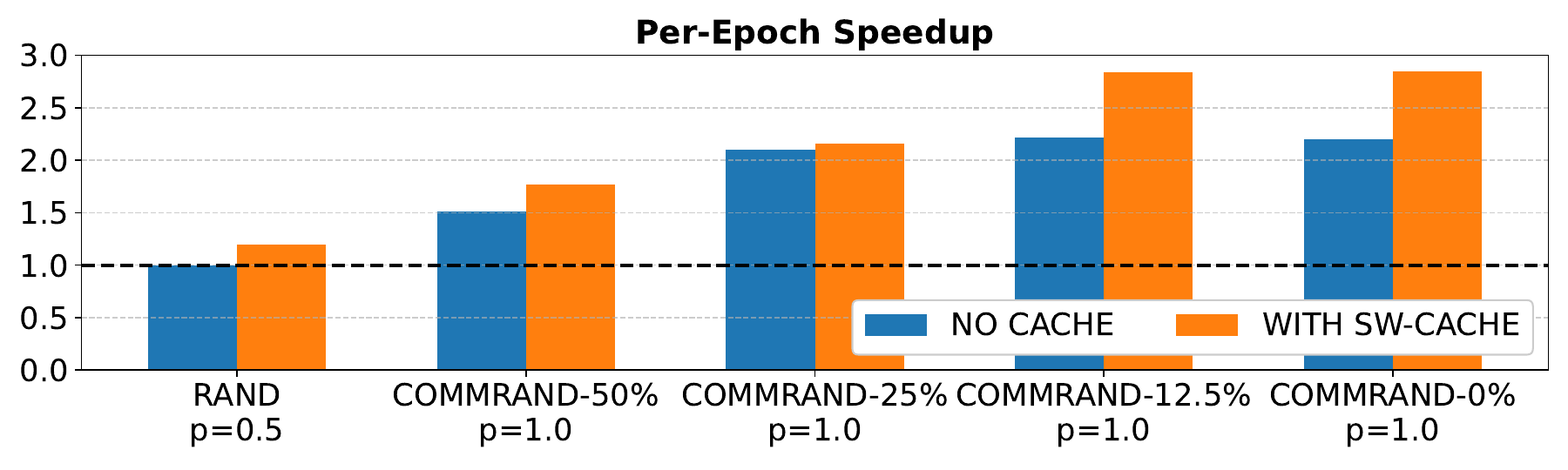}
\caption{{\bf Per-Epoch Training Speedups with a SW cache:}
{\em \sysname increases performance by reducing UVA transfers between CPU and GPU (for {\tt ogbn-papers100M}).}
}
\label{fig:sw-gpu-cache}
\end{figure}

As mentioned in Section~\ref{sec:methodology}, due to the large size of the {\tt \seqsplit{ogbn-papers100M}} dataset we could not fit the graph and features all on the GPU memory and, therefore, we performed mixed CPU$+$GPU training for this dataset.
Specifically, the graph and feature data are stored on the CPU memory while the model is on the GPU memory.
During neighborhood sampling, the relevant parts of the graph and feature data are moved from the CPU to the GPU using Unified Virtual Addressing (UVA)~\cite{uva, dgl-dataloader-api}.
The DGL framework provides an API for a software-managed cache that is resident in the GPU memory and allows reuse in feature accesses to reduce the amount of UVA transfers.
Figure~\ref{fig:sw-gpu-cache} compares the per-epoch speedups achieved with \sysname without any software caching (the results in Figure~\ref{fig:metrics-vs-knobs}) with per-epoch speedups in the presence of a software cache capable of storing 4M node features.
The result shows that the presence of a software cache allows \sysname to provide higher per-epoch training speedups.
The more community-biased versions of \sysname (i.e. {\tt \seqsplit{COMM-RAND-MIX-\{12.5\%,0\%\}}} $+$ $p=1.0$) offer greater speedups due to their ability to achieve higher reuse in the software cache.
With the software cache, the uniform random baseline experiences a miss rate of 35.46\% whereas the different \sysname versions from {\tt COMM-RAND-MIX-50\%} to {\tt COMM-RAND-MIX-0\%} experience miss rates of 20.99\%, 11.39\%, 6.22\%, and 6.21\% respectively.
Lastly, the software cache used in this experiment employs the LRU replacement policy~\cite{merlin-gpu-cache}.
Extending the software cache to use more graph-centric cache replacement policies~\cite{popt-hpca21,grasp-hpca20} could further boost \sysname's performance improvements and we will explore \sysname's effectiveness in caching node feature data~\cite{bgl-nsdi23,pagraph-socc20,global-neighborhood-sampling} as part of future work.

\subsubsection{Sensitivity to on-chip cache capacity} \label{subsec:l2-cache-sensitivity}

\begin{figure}[ht]
\centering
\includegraphics[keepaspectratio,width=\columnwidth]{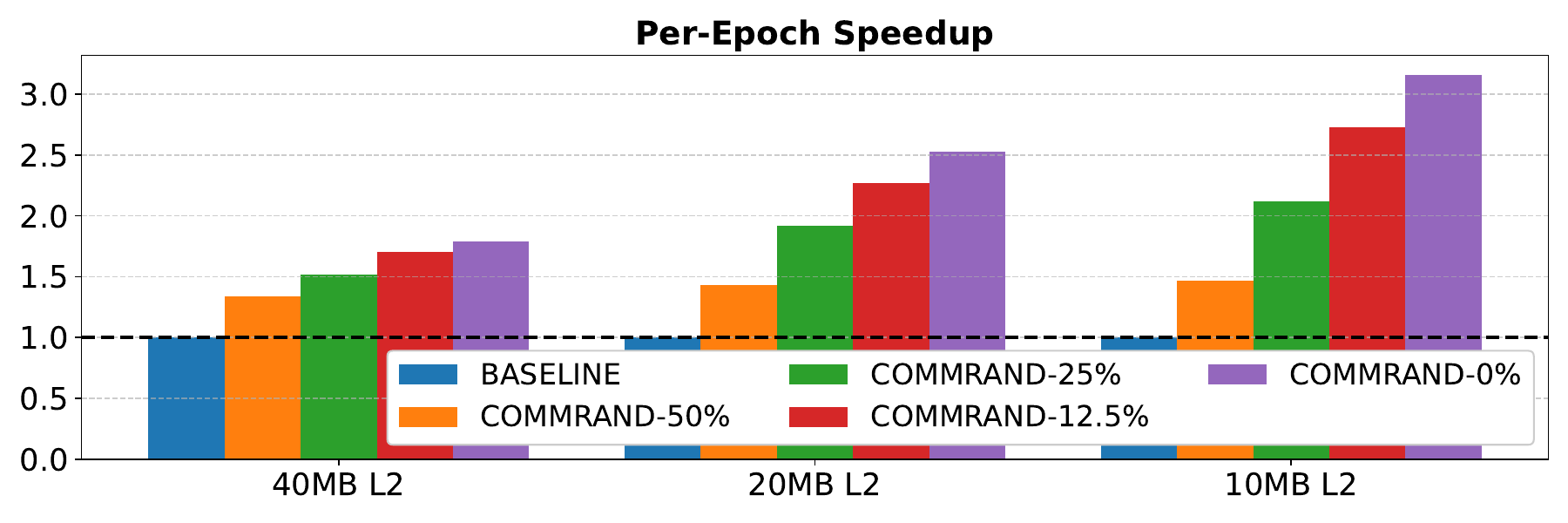}
\caption{{\bf Per-Epoch Speedups vs L2 cache capacity:}
{\em Results normalized to the baseline for each L2 configuration.}
}
\label{fig:l2-cache-sensitivity}
\end{figure}

The main reason for per-epoch training speedups with \sysname is due to reducing the working set size and improving on-chip cache reuse (Figure~\ref{fig:perepochtime-vs-featsize}).
To further explore \sysname's effectiveness a function of the L2 cache capacity, we used Multi-Instance GPU (MIG)~\cite{mig} to create GPU partitions with reduced L2 cache capacities.
An A100 GPU features a 40MB L2 cache and with MIG we created GPU partitions with 1/2 and 1/4 the number of SMs and memory capacity which have 20MB and 10MB L2 caches respectively.
Figure~\ref{fig:l2-cache-sensitivity} shows the per-epoch speedups with different \sysname configurations (with $p=1.0$) for the {\tt reddit} dataset across different L2 cache capacities.
The results show that as the L2 cache size decreases (or the ratio of the working set size relative to the cache capacity increases), we see higher per-epoch speedups across the different \sysname configurations.
While using MIG partitions is not a perfect experiment for a cache sensitivity study (because MIG also cuts the number of SMs and available memory bandwidth along with the L2 cache capacity), these results provide an estimate of \sysname's continued effectiveness as the ratio of the problem size relative to the cache capacity increases.

\subsubsection{Pre-processing overheads} \label{subsec:preprocessing-cost}
We assume that the graphs are already community-ordered for all runs (including the baseline) and, therefore, do not consider the pre-processing cost of reordering.
Furthermore, the overhead of graph reordering can be amortized over multiple epochs of GNN training (as well as inference).
Running RABBIT graph reordering on the {\tt reddit} dataset amounts to 0.78\% of the baseline's total training time.
Finally, \sysname only requires identifying the community membership of nodes and does not require the graph to be community-ordered which allows further reduction in pre-processing overheads.

\section{Related Work} \label{sec:related}

We discussed prior work on GNN mini-batching in Sections~\ref{sec:background} and ~\ref{subsec:prior-work-comparison}. In this section, we compare \sysname against general GNN training and inference optimizations.

Merkel et. al.~\cite{can-reordering-speedup-gnn-training-vldb24} performed an extensive study of the impact of different reordering techniques (including RABBIT) on GNN training.
However, their work largely focused on full-graph training (which has slower convergence vs mini-batch training; Section~\ref{sec:background}) and does not explore reordering's impact on training accuracy (which can be quite severe as shown in Figure~\ref{fig:rand-vs-norand-extreme:ogbn-papers100M}).
Similarly, Neugraph~\cite{neugraph-atc19} and ROC~\cite{roc-mlsys20} target the scalability challenges of full-graph GNN training.
SALIENT~\cite{salient-mlsys22} improves mini-batch GNN training and inference performance by pipelining the batch creation (on CPU) and data transfer steps with the training computations (on GPU) to maximize GPU utilization.
\sysname reduces the size of each batch's subgraph and, therefore, can help SALIENT's pipelining by saturating GPU utilization with fewer CPU processes.
Betty~\cite{betty-asplos23} proposed a partitioning strategy to sub-divide mini-batches into micro-batches in order to scale GNN training to larger graph sizes.
Betty maximizes memory savings by reducing redundancy across the micro-batches and \sysname's community-biased mini-batches are likely to help with finding low redundancy partitions (as shown in Figure~\ref{fig:biased-ngh-selection}).
There have been many GNN accelerator proposals~\cite{hygcn-hpca20,awb-gcn-micro20,grow-hpca23,igcn-micro21} but they primarily target GNN inference.
GNN inference typically operates on the entire graph and is invariant to the order in which nodes are processed (and therefore benefits from graph reordering optimization).
In contrast, mini-batch GNN training exhibits a more challenging access pattern where the number of nodes processed is a function of the training set of the graph and the need for randomization leads to a very dynamic reuse pattern that changes each training epoch.

%
%

\section{Conclusions} \label{sec:conclusions}
In this work, we highlight the tension in mini-batch construction between the need for randomization (for good accuracy and convergence) and the need to leverage the graph's community structure (for efficient per-epoch processing). We demonstrate that \sysname resolves this dilemma by introducing controllable structure-aware randomization that balances these competing requirements. By providing knobs that adjust the degree of community bias during mini-batching, \sysname enables practitioners to optimize the trade-off between convergence rate and computational efficiency, achieving up to 2.76$\times$ (1.8$\times$ on average) faster training while maintaining accuracy within 1.79\% points (0.42\% on average) of traditional approaches.

\bibliographystyle{ACM-Reference-Format}
\bibliography{example-paper}


\end{document}